\documentclass{article}

\usepackage{microtype}
\usepackage{graphicx}
\usepackage{subcaption}
\usepackage{booktabs}
\usepackage{hyperref}
\usepackage{adjustbox}

\usepackage[preprint]{icml2026}

\usepackage{amsmath,amssymb,mathtools}
\usepackage{amsthm}

\usepackage[capitalize,noabbrev]{cleveref}

\begin{document}

\twocolumn[
    \icmltitle{ROSA-Tuning: Enhancing Long-Context Modeling via Suffix Matching}

    \begin{icmlauthorlist}
      \icmlauthor{Yunao Zheng}{bupt}
      \icmlauthor{Xiaojie Wang}{bupt}
      \icmlauthor{Lei Ren}{liauto}
      \icmlauthor{Wei Chen}{liauto}
    \end{icmlauthorlist}
    
    \icmlaffiliation{bupt}{Beijing University of Posts and Telecommunications (BUPT), Beijing, China}
    \icmlaffiliation{liauto}{Li Auto Inc., Beijing, China}
    
    \icmlcorrespondingauthor{Yunao Zheng}{3027949562@qq.com}
    
    \vskip 0.3in
    
]

\printAffiliationsAndNotice{}

\begin{abstract}

Long-context capability and computational efficiency are among the central challenges facing today’s large language models. Existing efficient attention methods reduce computational complexity, but they typically suffer from a limited coverage of the model state. This paper proposes ROSA-Tuning, a retrieval-and-recall mechanism for enhancing the long-context modeling ability of pretrained models. Beyond the standard attention mechanism, ROSA-Tuning leverages in parallel a CPU-based ROSA (RWKV Online Suffix Automaton) retrieval module, which efficiently locates historical positions in long contexts that are relevant to the current query, and injects the retrieved information into the model state in a trainable manner; subsequent weighted fusion can then be handled by range-restricted attention. To enable end-to-end training, we employ the binary discretization strategy and the counterfactual gradient algorithm, and further optimize overall execution efficiency via an asynchronous CPU–GPU pipeline. Systematic evaluations on Qwen3-Base-1.7B show that ROSA-Tuning substantially restores the long-context modeling ability of windowed-attention models, achieving performance close to and in some cases matching global attention on benchmarks such as LongBench, while maintaining computational efficiency and GPU memory usage that are nearly comparable to windowed-attention methods, offering a new technical path for efficient long-context processing. The example code can be found at https://github.com/zyaaa-ux/ROSA-Tuning.

\end{abstract}

\section{Introduction}

Long-context capability and efficiency are among the core challenges faced by today’s large language models. This capability directly affects key behaviors such as long chain-of-thought reasoning and multi-turn dialogue consistency, and determines whether a model can reliably handle input sequences of tens of thousands of tokens or even longer in real-world applications. However, the high complexity of attention~\citep{vaswani2017attention} has become a critical bottleneck. Although optimization techniques such as Flash Attention~\citep{shah2024flashattention3} alleviate this issue to some extent via block-wise computation and improved memory access patterns, when processing contexts of tens of thousands of tokens or longer, the compute and GPU memory overheads still pose a severe challenge.

To reduce the cost of long-context processing, the community has mainly proposed three classes of approaches. Sparse attention~\citep{child2019sparse} reduces computation by selectively computing key query--key pairs, but faces an inherent trade-off between accuracy and efficiency: selecting too few tokens fails to model long-range dependencies adequately, while selecting too many tokens does not effectively reduce complexity. Linear attention~\citep{katharopoulos2020transformers} compresses the historical context into a fixed-size state, thereby achieving linear complexity, but its performance often degrades as the sequence length increases. Hybrid attention methods attempt to combine sparse connectivity with state compression to balance capacity and efficiency. According to the theory of \citet{katharopoulos2020transformers}, different efficient attention methods essentially restrict the effective state size that participates in computation. Consequently, these methods do not resolve the fundamental tension of attention mechanisms: a fixed-size state cannot cover extremely long contexts, whereas a variable-size state incurs computation that grows with sequence length.

We use the term “cannot cover” rather than “cannot handle”. The reason is that the true bottleneck is often not insufficient compression capacity, but rather limited state coverage. For example, RWKV-7~\citep{peng2025rwkv7} shows that a state of only 8096 dimensions can accommodate more than 1k tokens of context information, with a state information density up to 0.547 bit per dimension, demonstrating the feasibility of highly efficient compression. The practical issue is that the state space participating in computation does not include all historical information required by the current task, leading to failures in recalling critical details.

\begin{figure*}[htbp]
\centering
\includegraphics[width=\textwidth]{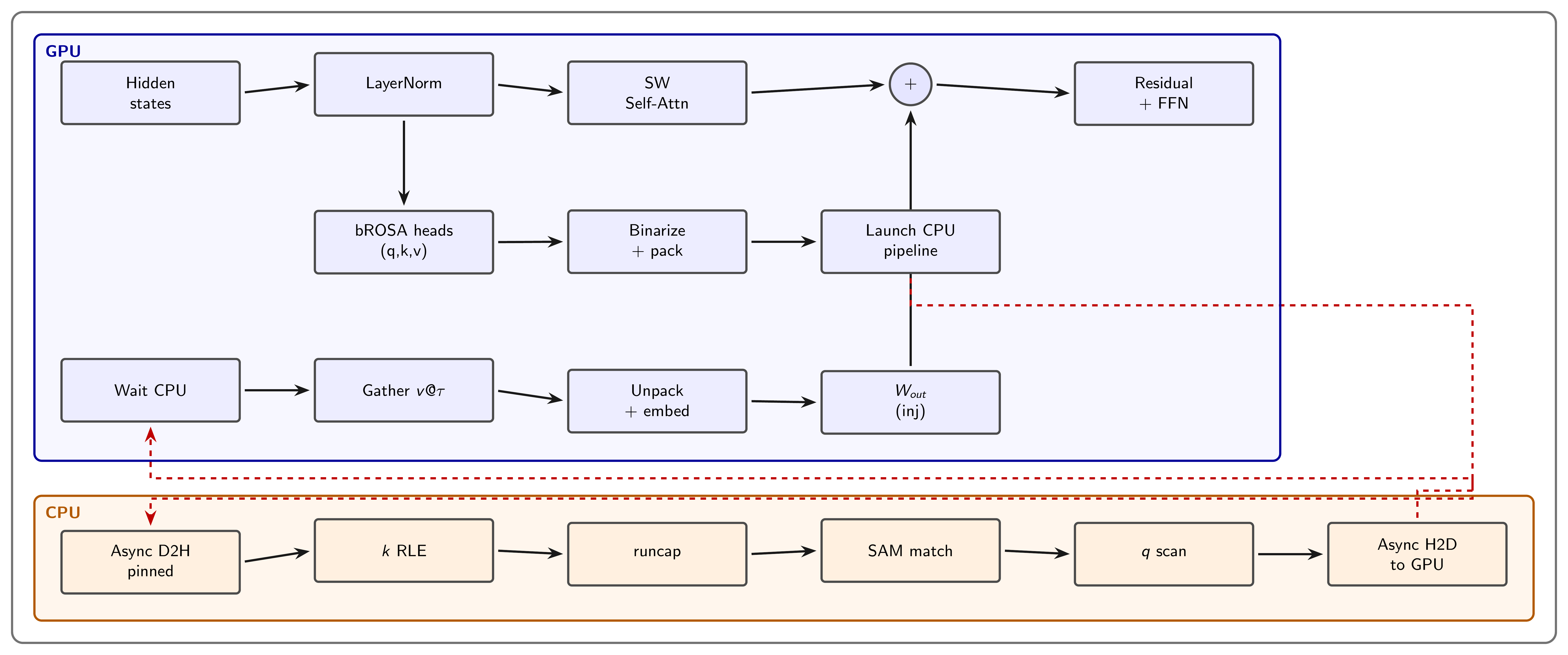}
\caption{ROSA-Tuning architecture}
\label{fig:rosa}
\end{figure*}

Based on these observations, we propose ROSA (RWKV Online Suffix Automaton)-Tuning, a method that introduces a retrieval-and-recall mechanism into pretrained models. As shown in Figure~\ref{fig:rosa}, ROSA-Tuning does not perform attention computation over all historical tokens. Instead, in parallel to attention, it introduces an efficient CPU-based retrieval process that identifies a small set of historical positions relevant to the current query from the long context, and injects the corresponding information into the model state in a trainable manner. The subsequent weighted fusion of information is still handled by the attention mechanism; therefore, the model can, in computation, use windowed attention to process input sequences of arbitrary length.

We systematically evaluate ROSA-Tuning on the Qwen3-Base~\citep{qwen2025qwen3} model, validating its effectiveness on both general-purpose tasks and long-context tasks, and compare its computational efficiency against the latest Flash Attention \citep{dao2022flashattention} implementation on an NVIDIA RTX 5090 GPU. The results show that, compared with the officially released sliding-window attention baseline, ROSA-Tuning substantially restores long-context modeling capability, with overall performance close to or matching the global-attention baseline. Moreover, when processing sequences of arbitrary length, models using ROSA-Tuning exhibit speed and GPU memory consumption that are almost identical to those of windowed attention. These results demonstrate that ROSA-Tuning effectively improves computational efficiency while maintaining long-context modeling capability.

\section{Background}

\subsection{Reducing computational complexity by shrinking the effective state size}

\citet{katharopoulos2020transformers} point out that, under a causal mask, the Transformer self-attention layer can be expressed in a recurrent form: its internal state is represented by a matrix accumulated from the outer products of historical key--value pairs, and the current output is obtained by multiplying the query vector with this state. Since practical implementations require softmax normalization over attention weights, the cost of accessing the state grows quadratically with the number of key--value pairs participating in computation, which constitutes the central computational bottleneck of self-attention in long-sequence settings.

From this perspective, the essential differences among efficient attention methods lie in how they restrict or reorganize the “effective state size accessible at each time step.” Concretely, global attention includes all historical key--value pairs in the state, yielding overall quadratic complexity. Windowed and sparse attention limit the number of key--value pairs that form the state, thereby controlling the per-step state access cost to $O(W)$ or $O(k)$, respectively, at the expense of reduced coverage of long-range dependencies. Linear attention compresses the entire history into a fixed-size state, achieving linear complexity, but suffers from issues such as error accumulation and thus cannot faithfully approximate softmax attention.

ROSA-Tuning offers a way to address the tension between efficiency and coverage: rather than further compressing the readable state inside attention, we introduce a low-cost recall module outside the attention mechanism. Without altering the structure of efficient attention, this module can retrieve relevant information online and inject it into the state representation, thereby effectively compensating for the limited long-range coverage of the attention mechanism.

\subsection{Efficient information retrieval with ROSA}

Attention computation can be decomposed into two stages: (i) generating the historical state readable at the current time step (i.e., a candidate set of key--value pairs), and (ii) performing weighted fusion over this state to produce the output. Under long-context settings, directly enlarging the readable state in stage (i) incurs substantial computational cost. A natural alternative is therefore to use an independent retrieval module to generate candidate key--value pairs, and then let attention perform the continuous weighted fusion in stage (ii). ROSA provides a suitable algorithmic foundation for this purpose.

A suffix automaton (SAM) is a compact string indexing structure that can represent, online, the set of all substrings of a sequence. The number of states has a linear upper bound for a sequence of length $n$ (no more than $2n-1$), and it supports amortized $O(1)$ state transitions and suffix-link jumps~\citep{blumer1985smallest}. This property allows SAM to perform retrieval operations of the form “jump from the current context to a relevant historical position” with extremely low computational overhead in streaming long-sequence processing. Building on this, ROSA ~\citep{peng_bo_2021_5196578} maintains hundreds to thousands of SAM instances in parallel to cover as many potential associations as possible, thereby obtaining comprehensive information.

ROSA-Tuning integrates the above retrieval mechanism with pretrained models. It first discretizes continuous hidden representations into a symbol sequence, and in parallel constructs ROSA-based retrieval structures outside the attention mechanism to quickly locate historical positions relevant to the current context. It then injects the retrieved information into the model in a trainable manner, while the subsequent weighted fusion is still carried out by local sliding-window attention. This design enhances the model’s ability to leverage long-range dependency information while preserving computational efficiency.

\section{Method}

\subsection{Overall Framework}

Consider a single-layer decoder block, whose input hidden states are denoted as
\[
\mathbf{H}\in\mathbb{R}^{B\times T\times C},
\]
where $B$ is the batch size, $T$ is the sequence length, and $C$ is the hidden dimension. Let windowed attention be $\mathrm{Attn}_W(\cdot)$, whose output is
\[
\mathbf{A}=\mathrm{Attn}_W(\mathrm{LN}(\mathbf{H}))\in\mathbb{R}^{B\times T\times C}.
\]

On top of this, ROSA-Tuning introduces an additional injection term
\[
\mathrm{inj}=\mathrm{ROSA}(\mathbf{H})\in\mathbb{R}^{B\times T\times C},
\]
where $\mathrm{inj}$ represents candidate features derived from global historical information. This injection term can be fused with the attention mechanism in two different ways.

\paragraph{post-attn (additive fusion)}
\begin{align}
\mathbf{A} &= \mathrm{Attn}_W(\mathrm{LN}(\mathbf{H})), \\
\mathbf{H}' &= \mathbf{H}+\mathbf{A}+\mathrm{inj}, \\
\mathbf{H}'' &= \mathbf{H}'+\mathrm{MLP}(\mathrm{LN}(\mathbf{H}')).
\end{align}

\paragraph{pre-attn (time mixing)}

Following RWKV's time-shift (time-mixing) formulation, we introduce a per-channel gating parameter $\boldsymbol{\alpha}=\sigma(\boldsymbol{\alpha}_0)\in(0,1)^C$, and linearly mix the hidden states with the ROSA injection term before feeding them into attention:
\begin{align}
\mathbf{M} &= (1-\boldsymbol{\alpha})\mathbf{H}+\boldsymbol{\alpha}\,\mathrm{inj}, \\
\mathbf{A} &= \mathrm{Attn}_W(\mathrm{LN}(\mathbf{M})), \\
\mathbf{H}' &= \mathbf{H}+\mathbf{A}, \\
\mathbf{H}'' &= \mathbf{H}'+\mathrm{MLP}(\mathrm{LN}(\mathbf{H}')).
\end{align}

The former allows ROSA and the attention module to execute in parallel; since the computational overhead of ROSA is significantly lower than that of attention, it can be regarded as an (approximately) ``zero-cost'' addition. The latter requires ROSA inference to be completed before attention computation; although it introduces extra overhead, it typically yields better performance in practice.

\subsection{Binary Discretization and Multi-Route Symbol Streams}

ROSA performs retrieval over discrete symbol sequences, so we first map continuous representations to symbol streams. To this end, we introduce adapter parameters at each layer that are decoupled from the backbone attention projections:
\begin{align}
\mathbf{U} &= \mathrm{LN}(\mathbf{H}), \\
\mathbf{Q}^{\mathrm{vec}} &= \mathbf{U}\mathbf{W}_q,\qquad
\mathbf{K}^{\mathrm{vec}} = \mathbf{U}\mathbf{W}_k,\qquad
\mathbf{V}^{\mathrm{vec}} = \mathbf{U}\mathbf{W}_v,
\end{align}
where $\mathbf{W}_q,\mathbf{W}_k,\mathbf{W}_v\in\mathbb{R}^{C\times C}$ are trainable parameters.

\textbf{Theorem 1} Viewing ROSA as a communication channel that transmits historical information into the state, under the same budget and limited noise, binarization is least likely to map identical content to different symbols, and when the distribution is close to uniform it also drives the collision probability down to the theoretical minimum.

The proof is provided in Appendix A. Following Theorem 1, we apply threshold binarization to each dimension:
\begin{align}
q^{\mathrm{bit}}_{b,t,c} &= \mathbb{I}\!\left[q^{\mathrm{vec}}_{b,t,c}>0\right],\\
k^{\mathrm{bit}}_{b,t,c} &= \mathbb{I}\!\left[k^{\mathrm{vec}}_{b,t,c}>0\right],\\
v^{\mathrm{bit}}_{b,t,c} &= \mathbb{I}\!\left[v^{\mathrm{vec}}_{b,t,c}>0\right].
\end{align}

We partition the $C$-dimensional features into routes of size $M$, so that the number of routes is
\[
\begin{aligned}
R &= \frac{C}{M}, 
c &\equiv (r,j),\quad r\in[0,R-1],\quad j\in[0,M-1].
\end{aligned}
\]
The symbol alphabet size for each route is
\[
K=2^M.
\]

We then pack the $M$ bits within each route into an integer symbol:
\begin{align}
a^{(q)}_{b,t,r} &= \sum_{j=0}^{M-1} q^{\mathrm{bit}}_{b,t,(r,j)}\,2^j, \label{eq:pack_q}\\
a^{(k)}_{b,t,r} &= \sum_{j=0}^{M-1} k^{\mathrm{bit}}_{b,t,(r,j)}\,2^j, \label{eq:pack_k}\\
a^{(v)}_{b,t,r} &= \sum_{j=0}^{M-1} v^{\mathrm{bit}}_{b,t,(r,j)}\,2^j. \label{eq:pack_v}
\end{align}

\subsection{ROSA Retrieval}

For any batch index $b$, route $r$, and time step $t$, ROSA produces a historical index
\[
\tau_{b,r,t}\in\{-1,0,\dots,t-1\},
\]
which specifies from which historical position $\tau_{b,r,t}$ to read the value symbol of this route; if no valid retrieval result exists, we set $\tau_{b,r,t}=-1$.

Let
\[
k_{1:t,r}=a^{(k)}_{b,1:t,r},\qquad q_{1:t,r}=a^{(q)}_{b,1:t,r}.
\]
ROSA maintains, online, a suffix automaton over the sequence $k_{1:t,r}$, and simultaneously maintains a matching state $s_{b,r,t}$ such that the substring represented by this state is the longest suffix of $q_{1:t,r}$ that has a match in $k_{1:t,r}$. Let $\mathrm{endpos}(s)$ denote the end position of the most recent occurrence of the matched substring in $k$. We then define the destination using successor-position retrieval as
\begin{align}
\tau_{b,r,t}
=
\begin{cases}
\mathrm{endpos}(s_{b,r,t})+1,
& \begin{aligned}
  &\text{if this position exists}\\
  &\text{and is } < t,
\end{aligned}\\
-1, & \text{otherwise.}
\end{cases}
\label{eq:dest_time}
\end{align}

Equation~\eqref{eq:dest_time} ensures that the read position is strictly from the past, thereby satisfying the causal constraint. Meanwhile, this mechanism implements the retrieval behavior in the symbol stream of ``jumping from the current context to a relevant historical continuation.''

\textbf{Theorem 2} \;
When the attention similarity degenerates into a $0$--$1$ match/mismatch indicator, and the normalization takes an extreme preference over matched items, the attention output degenerates into an equally weighted average of the values at all matched positions, resembling multi-route ROSA.

The proof is provided in Appendix B. Intuitively, attention is responsible for weighted fusion, while ROSA only needs to retrieve relevant information. Therefore, we quantize the attention score between two tokens to $1$ (relevant) or $0$ (irrelevant). Under this condition, ROSA can be viewed as a form of global attention without weighting capability; when combined with windowed attention, it can approximate global attention.

Moreover, natural-language symbol streams often exhibit substantial local repetition. To reduce redundant overhead from SAM updates and matching, ROSA applies adjacent folding (run-length encoding, RLE) to $a^{(k)}_{b,1:t,r}$ for each route: consecutive identical symbols are treated as a single run, and the SAM and matching state are updated only at run boundaries. In implementation, the SAM operates on the run-level symbol sequence and maintains an array of the starting time indices for each run. When a hit $\mathrm{endpos}$ is obtained at the run level, we map $\tau_{b,r,t}$ back to the original time axis as the start of the next run (if it exists and is $<t$); otherwise we set it to $-1$. This folding does not change the retrieval semantics, but can substantially shorten the effective sequence length and reduce the number of state updates.

\subsection{ROSA Output}

Given $\tau_{b,r,t}$, we define the validity mask
\begin{align}
m_{b,r,t} &= \mathbb{I}[\tau_{b,r,t}\ge 0].
\label{eq:mask}
\end{align}

For each route, we read the corresponding value symbol from the destination position and set it to zero when the destination is invalid:
\begin{align}
\tilde a^{(v)}_{b,t,r}
&\triangleq m_{b,r,t}\cdot a^{(v)}_{b,\tau_{b,r,t},r},
&\qquad \tilde a^{(v)}_{b,t,r}\in\{0,\dots,2^M-1\}
\label{eq:read_v_symbol}
\end{align}
When $\tau_{b,r,t}=-1$, we have $m_{b,r,t}=0$, and thus $\tilde a^{(v)}_{b,t,r}=0$.

Next, we unpack the integer value symbol read from each route into binary bits. Let the dimension index $c$ be in one-to-one correspondence with the tuple $(r,j)$ (i.e., $c\leftrightarrow (r,j)$), where $j=0,\dots,M-1$ denotes the bit position. Then,
\begin{align}
b_{b,t,(r,j)}
&=
\left(\left\lfloor \frac{\tilde a^{(v)}_{b,t,r}}{2^j}\right\rfloor \bmod 2\right),
\qquad j=0,\dots,M-1
\label{eq:unpack}
\end{align}

For each continuous dimension $c$, we introduce two sets of learnable parameters $e_{0,c}$ and $e_{1,c}$, and define
\begin{align}
\Delta_c &= e_{1,c}-e_{0,c}.
\label{eq:delta}
\end{align}
We then define the continuous injection base vector as
\begin{align}
y_{b,t,c}
&=
m_{b,r,t}\,\Big(e_{0,c}+\Delta_c\,b_{b,t,c}\Big),
\label{eq:y}
\end{align}
where $b_{b,t,c}$ denotes the bit corresponding to dimension $c$ (i.e., $b_{b,t,c}\equiv b_{b,t,(r,j)}$), and the mask $m_{b,r,t}$ is broadcast according to the route to which the dimension belongs.

Finally, we obtain the injection vector via an output projection:
\begin{align}
\mathrm{inj}_{b,t,:} &= \mathbf{W}_{\mathrm{out}}\,y_{b,t,:},\qquad
\mathbf{W}_{\mathrm{out}}\in\mathbb{R}^{C\times C}.
\label{eq:inj}
\end{align}

With initialization $e_0=e_1=\mathbf{0}$ and $\mathbf{W}_{\mathrm{out}}=\mathbf{I}$, we have $\mathrm{inj}\equiv\mathbf{0}$ for any input.
Therefore, ROSA-Tuning can be inserted without changing the initial behavior of the pretrained model, and the recall pathway is gradually activated during training.

\subsection{Backpropagation}
\label{sec:cfgrad}

The forward path of ROSA-Tuning contains two classes of discrete operators. The first is the hard-threshold binarization $\mathbb{I}[x>0]$ and the subsequent bit packing (Equations~\eqref{eq:pack_q}--\eqref{eq:pack_v}); the second is the deterministic retrieval operator based on the suffix automaton (Equation~\eqref{eq:dest_time}). As a result, the injection term $\mathrm{inj}$ is a piecewise-constant function of $\mathbf{Q}^{\mathrm{vec}},\mathbf{K}^{\mathrm{vec}},\mathbf{V}^{\mathrm{vec}}$: small perturbations in the continuous space are often insufficient to change the binarization outcomes or the retrieval destination $\tau$, making the gradient along the true discrete path almost everywhere $0$. If one directly applies the straight-through estimator (STE;~\citealp{bengio2013ste}) to forcibly assign gradients to the threshold function, STE fails to reflect the structured dependency of ``bits $\rightarrow$ retrieval destination $\tau$ $\rightarrow$ read-out values,'' causing the gradient direction to decouple from the effect of the true discrete decisions and leading to unstable or even divergent training in practice.

To address this, we adopt a counterfactual gradient strategy by treating each query/key bit as a discrete decision switch. For a given bit $b$, we construct two counterfactual branches---``force $b=0$'' and ``force $b=1$''---and perform one retrieval update on the same historical state to obtain the destinations and read-out results for the two branches. In this way, the influence of the bit on the loss can be characterized by the difference between the two counterfactual read-outs. This approach yields accurate gradients without random sampling and explicitly aligns with ROSA's retrieval structure.

Let the training loss be $\mathcal{L}$. From Equation~\eqref{eq:inj}, we define
\[
\mathbf{G}^{\mathrm{inj}}_{b,t,:} \triangleq \frac{\partial \mathcal{L}}{\partial \mathrm{inj}_{b,t,:}},\qquad
\mathbf{G}^{y}_{b,t,:} \triangleq \frac{\partial \mathcal{L}}{\partial y_{b,t,:}}
= \mathbf{W}_{\mathrm{out}}^{\top}\mathbf{G}^{\mathrm{inj}}_{b,t,:}.
\]
We further define the dimension-wise weighted residual
\begin{equation}
\theta_{b,t,c} \triangleq G^y_{b,t,c}\,\Delta_c,
\label{eq:theta_def}
\end{equation}
where $\Delta_c=e_{1,c}-e_{0,c}$ (see Equation~\eqref{eq:delta}).

\textbf{Gradients w.r.t.\ $(e_0,e_1,\mathbf{W}_{\mathrm{out}})$ (directly differentiable).}
From Equations~\eqref{eq:y}--\eqref{eq:inj}, the gradients of $(e_0,e_1,\mathbf{W}_{\mathrm{out}})$ can be computed directly via the standard chain rule.
Closed-form expressions and the full derivation are provided in Appendix C.3 and Equation~\eqref{eq:grad_wout}).

\textbf{Gradients w.r.t.\ $\mathbf{V}^{\mathrm{vec}}$ (destination-scatter aggregation).}
To make the value branch differentiable, in backpropagation we use the continuous surrogate
$P^{(v)}=\sigma(\mathbf{V}^{\mathrm{vec}})$
to approximate the binary values; the local derivative of each bit is given by $\sigma'(x)=\sigma(x)(1-\sigma(x))$. Since in the forward pass the read-out at each time step $t$ comes from destination $\tau_{b,r,t}$, in the backward pass the gradients propagate along this ``read pointer'' and accumulate at the same destination. Concretely, the gradient of $v^{\mathrm{vec}}$ can be written in a scatter-aggregation form over the retrieval destination $\tau$ (see Appendix C.4 for the derivation):
\begin{equation}
\frac{\partial \mathcal{L}}{\partial v^{\mathrm{vec}}_{b,\tau,c}}
=
\sigma'\!\big(v^{\mathrm{vec}}_{b,\tau,c}\big)
\sum_{t=0}^{T-1}
\theta_{b,t,c}\,
\mathbb{I}\!\big[\tau_{b,r(c),t}=\tau\big],
\label{eq:grad_v_main}
\end{equation}
where $r(c)$ denotes the route to which dimension $c$ belongs ($c\leftrightarrow(r,j)$).

\textbf{Gradients w.r.t.\ $\mathbf{Q}^{\mathrm{vec}}$ (bitwise counterfactual differencing).}
For any time step $t$, route $r$, and bit $j$ within this route,
we precompute the counterfactual retrieval destinations
$\tau^{(0)}_{b,t,r,j}$ and $\tau^{(1)}_{b,t,r,j}$ when the bit is forced to $0$ or $1$, respectively (see Appendix C.5 for details).
For all bit dimensions $m\in\{0,\dots,M-1\}$ within the same route, we define the counterfactual difference in the value surrogate read-out as
\[
\delta P^{(v)}_{b,t,r,m}(j)
\triangleq
P^{(v)}_{b,\tau^{(1)}_{b,t,r,j},(r,m)}
-
P^{(v)}_{b,\tau^{(0)}_{b,t,r,j},(r,m)}.
\]
Then,
\begin{equation}
\frac{\partial \mathcal{L}}{\partial q^{\mathrm{vec}}_{b,t,(r,j)}}
=
\sigma'\!\big(q^{\mathrm{vec}}_{b,t,(r,j)}\big)\,
\sum_{m=0}^{M-1}
\theta_{b,t,(r,m)}\,
\delta P^{(v)}_{b,t,r,m}(j).
\label{eq:grad_q_main}
\end{equation}

\textbf{Gradients w.r.t.\ $\mathbf{K}^{\mathrm{vec}}$ (counterfactual differencing with run-level aggregation).}

Since adjacent folding is applied to the key symbol sequence, the suffix automaton operates on the run-level sequence.
To avoid the high cost of explicitly flipping key bits, we introduce a differentiable surrogate at the run level.
Specifically, for each run $\ell$, route $r$, and bit $j$, we define a continuous gate at the run start
\[
u_{b,\ell,r,j} \triangleq \sigma\!\big(k^{\mathrm{vec}}_{b,\mathrm{start}(\ell),(r,j)}\big).
\]
Meanwhile, let $r\_idx^{(0)}_{b,t,r,j}$ and $r\_idx^{(1)}_{b,t,r,j}$ denote the run-level destination indices corresponding to the two counterfactual branches obtained by forcing the $j$-th query bit to $0/1$ (with all other bits unchanged). We then obtain the following run-level gradient (see Appendix C.6 for the derivation):
\begin{equation}
\frac{\partial \mathcal{L}}{\partial k^{\mathrm{vec}}_{b,\mathrm{start}(\ell),(r,j)}}
=
\sigma'\!\big(k^{\mathrm{vec}}_{b,\mathrm{start}(\ell),(r,j)}\big)\,
\Big(U^{(1)}_{b,\ell,r,j}-U^{(0)}_{b,\ell,r,j}\Big).
\label{eq:grad_k_main}
\end{equation}
Finally, we scatter this gradient back to the original time positions via the run-start indices, while ignoring higher-order effects of within-run positions on the folding boundaries.

\textbf{Gradients w.r.t.\ $(\mathbf{W}_q,\mathbf{W}_k,\mathbf{W}_v)$ and the gating parameters.}
After obtaining
$\partial \mathcal{L}/\partial \mathbf{Q}^{\mathrm{vec}}$,
$\partial \mathcal{L}/\partial \mathbf{K}^{\mathrm{vec}}$, and
$\partial \mathcal{L}/\partial \mathbf{V}^{\mathrm{vec}}$,
the gradients of the projection matrices can be computed directly via the standard backpropagation of linear layers.
In addition, the mixing gate $\boldsymbol{\alpha}$ in pre-attention remains differentiable throughout, and its gradient can likewise be computed by the chain rule; the relevant formulas and derivations are consolidated in the last subsection of Appendix~\ref{app:cfgrad}.

\section{Implementation}
\label{sec:impl}

This section introduces two key engineering optimizations for ROSA-Tuning, including the execution-order design between the CPU and GPU and optimization strategies for parallel retrieval. These optimizations do not change the algorithmic definition of ROSA-Tuning; they are solely used to reduce memory footprint and improve execution efficiency.

\subsection{Parallel Retrieval}
\label{sec:impl_parallel}

In ROSA-Tuning, the retrieval procedure can be decomposed into a large number of mutually independent subtasks and executed in parallel on the CPU. Concretely, we partition the hidden dimension into routes of size $M$, with the number of routes given by $R=C/M$. At each $(b,r)$ position, we independently maintain the corresponding symbol stream and retrieval structure, and output the destination pointers $\tau_{b,r,1:T}$ along with auxiliary tensors required for backpropagation.

Since there are no data dependencies across different $(b,r)$ pairs, the retrieval process can be parallelized at the granularity of $B\times R$, thereby substantially improving CPU-side throughput.

\subsection{CPU--GPU Execution Order}
\label{sec:impl_pipeline}

A single forward pass of ROSA-Tuning consists of the following steps: discretization and packing on the GPU, retrieval and construction of the counterfactual tables on the CPU, and result transfer back to the GPU followed by injection fusion. The overall procedure is as follows:
\begin{enumerate}
  \item \textbf{GPU compute stage}: compute $\mathbf{U}=\mathrm{LN}(\mathbf{H})$ and $\mathbf{Q}^{\mathrm{vec}}$, $\mathbf{K}^{\mathrm{vec}}$, $\mathbf{V}^{\mathrm{vec}}$, then perform threshold binarization and pack the results along the route dimension into integer symbols
  \[
  a^{(q)}, a^{(k)}, a^{(v)} \in \{0,\dots,2^M-1\}^{B\times T\times R}.
  \]
  \item \textbf{Asynchronous GPU$\rightarrow$CPU transfer}: in a dedicated copy stream, asynchronously transfer $a^{(q)}$ and $a^{(k)}$ to a host-pinned buffer, and record an event $E_{\text{copy}}$ on the copy stream. The CPU waits only for this event, without introducing synchronization with the default stream.
  \item \textbf{CPU retrieval stage}: perform symbolic retrieval and output the destination pointers $\tau_{b,r,t}$, run-start indices, the mapping between queries and runs, and the per-bit counterfactual candidate tables.
  \item \textbf{CPU$\rightarrow$GPU transfer and fusion}: asynchronously transfer the above retrieval results back to the GPU. The GPU reads the corresponding $a^{(v)}$ according to the destination pointers to construct the injection term $\mathrm{inj}$, and then completes the fusion computation.
\end{enumerate}

In the post-attn mode, the CPU-side retrieval can run in parallel with the GPU-side attention computation. Concretely, after launching the asynchronous device-to-host (D2H) transfer, the GPU immediately executes sliding-window attention
\[
\mathbf{A}=\mathrm{Attn}_W(\mathbf{U}).
\]
After attention finishes, the GPU waits for the CPU to return the retrieval results, constructs $\mathrm{inj}$, and finally performs additive fusion
\[
\mathbf{H}'=\mathbf{H}+\mathbf{A}+\mathrm{inj}.
\]
This execution order hides most of the CPU computation cost under the attention computation, making the additional end-to-end overhead close to zero.

In contrast, in the \textbf{pre-attn} mode, $\mathrm{inj}$ must be obtained first in order to construct the mixed input
\[
\mathbf{M}=(1-\alpha)\mathbf{H}+\alpha\,\mathrm{inj}
\]
and feed it into the attention module. Therefore, within the same layer, this mode is harder to overlap effectively with attention computation, and its performance is more sensitive to implementation constants and system bandwidth. Nevertheless, in practice this mode often yields stronger model quality, but requires bandwidth optimization and constant-factor optimization of retrieval to control the throughput degradation during training.

\section{Experiments}
\label{sec:exp}

We evaluate ROSA-Tuning based on Qwen3-Base-1.7B from three aspects: general capabilities, long-context modeling capability, and computational efficiency, and compare it against global-attention and windowed-attention baselines. In principle, ROSA-Tuning is applicable to any model that does not maintain global state access (e.g., windowed/sparse/linear attention). We choose windowed attention as the primary baseline because Qwen3-Base-1.7B provides both full-attention and windowed-attention variants, allowing us to apply ROSA-Tuning to the windowed-attention model and directly compare against both baselines under a unified architecture. Additional theoretical validation and hyperparameter-related experimental results are provided in Appendix D.

\subsection{Pretraining Setup}

In this section, the window size is set to 2048 throughout, $M$ in ROSA is set to 4, and the fusion mode with attention is post-attn.

The training pipeline consists of three stages: an initial adapter warm-up, long-context continued pretraining, and supervised fine-tuning.  
In the initial stage, we use approximately 4B tokens and train only the newly introduced ROSA-related parameters while keeping the backbone model parameters frozen.  
In the long-context continued pretraining stage, we unfreeze all parameters and continue training on approximately 26B tokens, with the backbone learning rate decayed from $5\times10^{-6}$ to $1\times10^{-6}$ via a cosine schedule.  
In the supervised fine-tuning stage, we train on approximately 7B tokens.

Due to limited compute resources, the model is not trained to full convergence, but the training scale is sufficient to validate the effectiveness of ROSA-Tuning.

\subsection{General Capability Evaluation}
\label{sec:exp:lmeval}

General capability evaluation is conducted using the lm-eval-harness~\cite{biderman2024lmeval} framework, covering six representative tasks spanning language understanding and commonsense reasoning. Table~\ref{tab:lmeval} shows that after ROSA-Tuning with substantial training data, the metrics exhibit only minor fluctuations, indicating that ROSA-Tuning has almost no impact on general capabilities.

\begin{table*}[t]
\centering
\small
\setlength{\tabcolsep}{4pt}
\caption{lm-eval results}
\begin{adjustbox}{max width=\linewidth}
\begin{tabular}{lccccccc}
\toprule
\textbf{Model} & \textbf{HellaSwag} & \textbf{LAMBADA-OAI} & \textbf{MMLU} & \textbf{PIQA} & \textbf{SciQ} & \textbf{Winogrande} & \textbf{AVG} \\
\midrule
Qwen3-1.7B (Global-Attn) & 0.6648 & 0.6295 & 0.6048 & 0.7568 & 0.9590 & 0.6448 & \textbf{0.7100} \\
Qwen3-1.7B (Window-Attn + ROSA) & 0.6558 & 0.6256 & 0.6033 & 0.7519 & 0.9540 & 0.6393 & 0.7050 \\
\bottomrule
\end{tabular}
\end{adjustbox}
\label{tab:lmeval}
\end{table*}

\subsection{Long-Context Evaluation}
\label{sec:exp:long}

As shown in Table~\ref{tab:longbench}, on long-sequence tasks~\cite{bai2024longbench}, the windowed-attention model after ROSA-Tuning significantly outperforms the original windowed-attention baseline, and approaches or even matches the global-attention model on most tasks. This suggests that ROSA can effectively retrieve key information from the historical context and incorporate it into the current-state computation, thereby substantially restoring the long-context modeling capability of windowed-attention models.

\begin{table*}[t]
\centering
\small
\setlength{\tabcolsep}{4pt}
\caption{LongBench results}
\begin{adjustbox}{max width=\linewidth}
\begin{tabular}{lccccccc}
\toprule
\textbf{Model} & \textbf{SAMSum} & \textbf{TriviaQA} & \textbf{MultiNews} & \textbf{TREC} & \textbf{GovReport} & \textbf{NIAH-32k} & \textbf{AVG} \\
\midrule
Qwen3-1.7B (Global-Attn) & \textbf{42.04} & \textbf{86.20} & 23.23 & \textbf{72.67} & \textbf{31.11} & \textbf{100.00} & \textbf{59.21} \\
Qwen3-1.7B (Window-Attn, $W{=}2048$) & 32.51 & 61.56 & 10.43 & 52.67 & 13.08 & 6.20 & 29.41 \\
Qwen3-1.7B (Window-Attn + ROSA) & 40.53 & 84.34 & \textbf{23.76} & 68.00 & 26.19 & \textbf{100.00} & 57.14 \\
\bottomrule
\end{tabular}
\end{adjustbox}
\label{tab:longbench}
\end{table*}

\subsection{Efficiency Analysis}
\label{sec:exp:efficiency}

ROSA-Tuning aims to introduce a low-cost recall-and-retrieval pathway without altering the core computation of windowed attention, enabling the model to process inputs of arbitrary length in a windowed-attention form. In terms of computational complexity, global attention has complexity $O(T^2)$, while windowed attention has complexity $O(TW)$; Window-Attn + ROSA maintains $O(TW)$ complexity on the GPU side, and the additional ROSA retrieval is executed primarily on the CPU side with approximately $O(T)$ complexity. Meanwhile, ROSA's states are stored mainly in CPU memory, so the GPU memory footprint remains essentially the same as that of the original windowed-attention model.

\begin{figure}[htbp]
  \centering
  \includegraphics[width=0.45\textwidth]{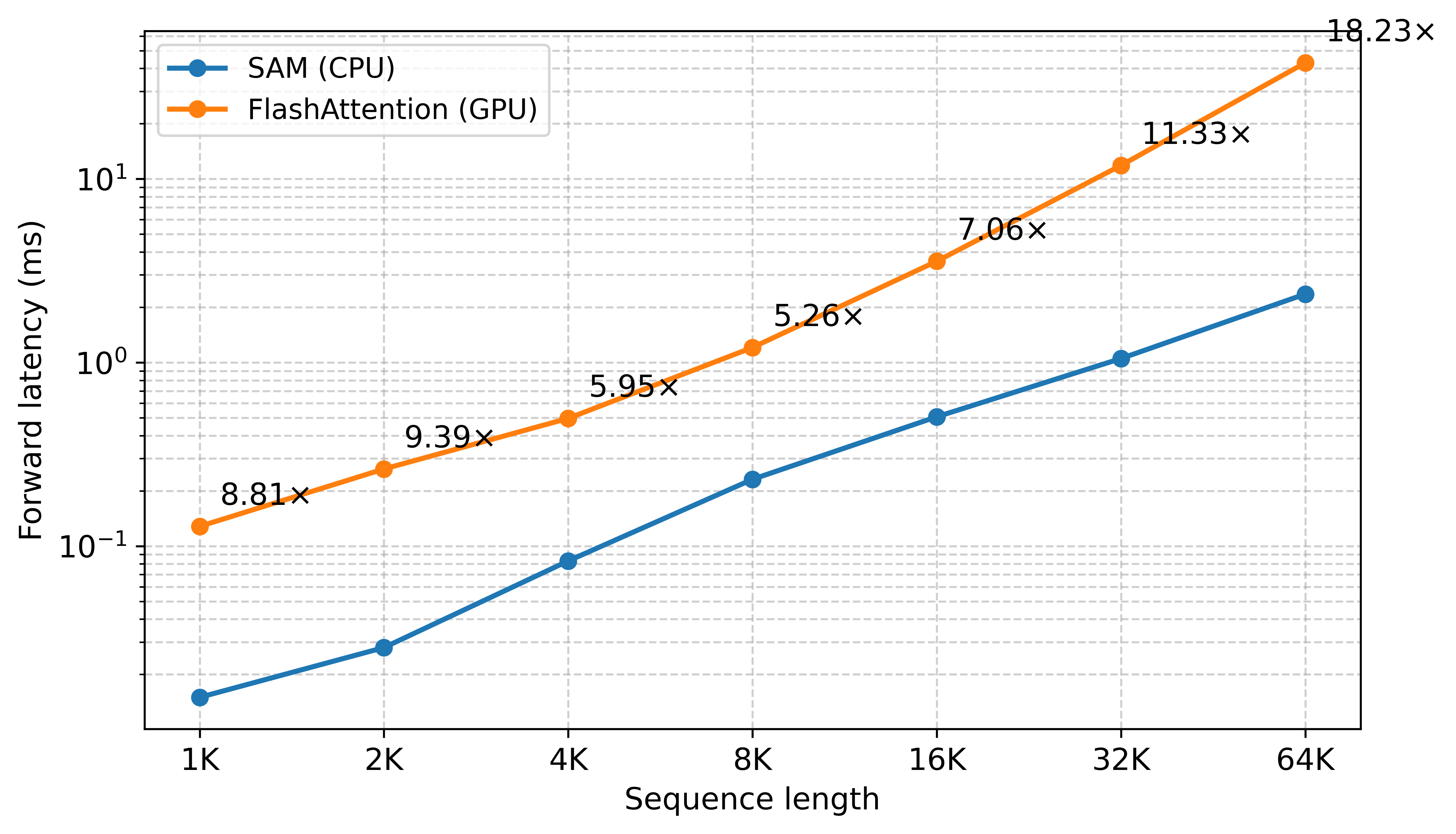}
  \caption{Runtime comparison.}
  \label{fig:time}
\end{figure}

At the implementation level, ROSA maintains an independent SAM and matching state for each batch and each route, and executes them in parallel on a multi-core CPU. Since end-to-end throughput is highly dependent on hardware configurations and implementation details, it is difficult to provide absolute speed numbers that are stable across platforms. Therefore, we compare the compute overhead of a single SAM on a single CPU core (ROSA can be viewed as executing multiple SAMs in parallel across cores, with wall-clock time comparable to that of a single SAM) with that of a very small attention kernel (1024 dimensions, FlashAttention implementation) on an NVIDIA RTX 5090 GPU. As shown in Figure~\ref{fig:time}, even for such a small-scale FlashAttention kernel, its compute cost is still substantially higher than that of a single SAM. Therefore, under most configurations, the additional overhead introduced by ROSA is almost entirely hidden by the attention computation; in particular, under the post-attention pipelined parallel mode, ROSA's compute overhead is essentially negligible.

\section{Related Work and Future Plans}

Recently, the Engram method proposed by DeepSeek~\citep{cheng2026conditionalmemoryscalablelookup} has attracted wide attention. Engram retrieves via input suffixes from a sparse table and injects the retrieved pretrained knowledge into the backbone model. This idea is somewhat similar to ROSA-Tuning. The key difference is that ROSA-Tuning retrieves using input suffixes from suffixes within the historical context, and injects the obtained historical information into the backbone network. Since the two methods retrieve from different sources, they differ in concrete implementations such as discretization strategies and training procedures; nonetheless, the overall idea is consistent. Notably, our work predates Engram.

Furthermore, Engram and ROSA-Tuning are complementary and can be combined to retrieve both historical context information and pretrained knowledge bases. We have begun related experiments, and the implementation details and experimental results will be released in the future.

\section{Conclusion}

This paper addresses the tension between state coverage and computational efficiency in long-context processing for large language models by proposing ROSA-Tuning. The core idea is to decouple retrieval from attention: a CPU-side ROSA module running in parallel efficiently identifies historically relevant information and injects it into windowed-attention computation in a trainable manner, thereby achieving effective coverage over contexts of arbitrary length while maintaining $O(TW)$ complexity and essentially the same GPU memory footprint as windowed attention. We employ the binary discretization strategy and the counterfactual gradient algorithm to enable end-to-end training, and further optimize execution efficiency via an asynchronous pipeline. Systematic experiments on Qwen3-Base show that the proposed method substantially restores long-context modeling performance while preserving general capabilities, approaching the global-attention baseline. The MQAR task further validates its retrieval alignment ability, providing a practical solution for efficient long-sequence processing in pretrained models.

\section{Acknowledgements}

During the research and writing of this thesis, we received valuable support and assistance from many individuals and institutions, to whom sincere gratitude is hereby expressed.

First and foremost, we would like to thank the advisor for their careful guidance and continuous support throughout the research process. Appreciation is also extended to the university and the affiliated school for providing a supportive research environment and academic atmosphere. We gratefully acknowledges Li Auto for providing computational resources, which enabled the efficient and stable execution of experiments. This work is based on the reproduction and application of Bo’s method; Bo offered insightful and forward-looking guidance on methodological understanding, experimental details, and research ideas (e.g., discretization strategies and inverse-gradient thinking), which played a crucial role in this study. Finally, sincere thanks are given to the researchers and contributors of the RWKV community for their valuable discussions and for helping identify and correct key issues during the writing process, significantly improving the quality and accuracy of this thesis.

\bibliography{refs}
\bibliographystyle{icml2026}

\newpage
\appendix
\onecolumn

\section{Analysis of Hit Stability and Spurious Collision Rate for Binary Discretization}
\label{app:thm1}

This appendix provides a formal proof of Theorem~1. We view ROSA's discrete retrieval process as a communication channel that transmits information from the historical state to the current state. The discretization scheme determines two properties of this channel: (i) \emph{hit stability}, i.e., whether the same underlying semantics are mapped to the same discrete symbol under different views; and (ii) the \emph{spurious collision rate}, i.e., whether semantically irrelevant historical positions are mistakenly retrieved due to symbol collisions.

\subsection{Formal Setup and Metric Definitions}

We analyze the single-layer, single-route case. One route corresponds to $M$ dimensions of the hidden vector
(see the main text $c\equiv (r,j)$, $j\in\{0,\ldots,M-1\}$).
Let the $M$-dimensional continuous representation be the random vector
\[
\mathbf{X}=(X_0,\ldots,X_{M-1})\in\mathbb{R}^M .
\]

For each dimension $j$, we use the same $L$-level threshold quantizer $Q_L:\mathbb{R}\to\{0,1,\ldots,L-1\}$,
defined by thresholds $-\infty=t_0<t_1<\cdots<t_{L-1}<t_L=+\infty$:
when $x\in(t_\ell,t_{\ell+1}]$, we set $Q_L(x)=\ell$.
(If different dimensions use different threshold sets, the derivations below remain unchanged in form by replacing $t_i$ with $t^{(j)}_i$.)

The same semantics yields two perturbed observations under the query view and the key view:
\begin{equation}
\begin{aligned}
X^{(q)}_j &= X_j + \varepsilon_{q,j},\\
X^{(k)}_j &= X_j + \varepsilon_{k,j},\\
|\varepsilon_{q,j}| &\le \delta,\quad
|\varepsilon_{k,j}| \le \delta\qquad \text{a.s.}\ \ \forall j\in\{0,\ldots,M-1\}.
\end{aligned}
\label{eq:bounded-noise}
\end{equation}

Define the per-dimension quantized digits:
\[
D^{(q)}_j = Q_L(X^{(q)}_j),\qquad
D^{(k)}_j = Q_L(X^{(k)}_j),
\]
and pack the $M$ digits into a single base-$L$ symbol (consistent with the binary packing in the main text):
\begin{equation}
Z^{(q)}=\sum_{j=0}^{M-1} D^{(q)}_j\,L^j,\qquad
Z^{(k)}=\sum_{j=0}^{M-1} D^{(k)}_j\,L^j,
\label{eq:pack-L}
\end{equation}
so the vocabulary size is
\begin{equation}
K=L^M.
\label{eq:K}
\end{equation}

We define hit stability as the probability that the two-view symbols agree:
\begin{equation}
\mathrm{Stab}(L,M)\triangleq
\Pr\!\big[ Z^{(q)} = Z^{(k)} \big].
\label{eq:stability-def}
\end{equation}

\subsection{Hit Stability Analysis}

We first analyze, under a fixed vocabulary budget $K$, how the discretization level $L$ affects hit stability.

\subsubsection{Lemma~\label{lem:digit-flip}: A sufficient condition for per-dimension digit agreement}
For any dimension $j$, if
\[
\min_{1\le i\le L-1} |X_j - t_i| > \delta,
\]
then under model~\eqref{eq:bounded-noise} we must have
\[
Q_L(X^{(q)}_j)=Q_L(X^{(k)}_j).
\]

\paragraph{Proof:}
When the distance from $X_j$ to every threshold $t_i$ exceeds $\delta$,
the two perturbed observations $X^{(q)}_j$ and $X^{(k)}_j$ must lie in the same quantization interval $(t_\ell,t_{\ell+1}]$,
and hence yield identical quantized outputs.

\subsubsection{Lemma~\label{lem:one-digit-bound}: An upper bound on per-dimension digit mismatch probability}
Assume that each $X_j$ has probability density function $f_{X_j}$ satisfying
\[
\sup_x f_{X_j}(x)\le f_{\max}\qquad \forall j\in\{0,\ldots,M-1\}.
\]
Then for any $j$,
\begin{equation}
\Pr\!\big[ Q_L(X^{(q)}_j)\neq Q_L(X^{(k)}_j) \big]
\le 2\delta f_{\max}(L-1).
\label{eq:one-digit-mismatch}
\end{equation}

\paragraph{Proof:}
By Lemma~\ref{lem:digit-flip}, a mismatch can occur only when $X_j$ falls within a $\delta$-neighborhood of some threshold $t_i$,
i.e., the event $\{|X_j-t_i|\le\delta\}$.
Therefore,
\[
\begin{aligned}
\Pr\!\big[Q_L(X^{(q)}_j)\neq Q_L(X^{(k)}_j)\big]
&\le \sum_{i=1}^{L-1}\Pr(|X_j-t_i|\le\delta) \\
&\le (L-1)\cdot 2\delta f_{\max},
\end{aligned}
\]
where the last inequality uses the density upper bound.

\subsubsection{Lemma~\label{lem:packed-stab}: A lower bound on stability after packing}
Under the above conditions,
\begin{equation}
\Pr[Z^{(q)}\neq Z^{(k)}]
\le \sum_{j=0}^{M-1}\Pr\!\big[ D^{(q)}_j\neq D^{(k)}_j\big]
\le 2\delta f_{\max}M(L-1).
\end{equation}

\begin{equation}
\mathrm{Stab}(L,M)
\ge 1-2\delta f_{\max}M(L-1).
\label{eq:stab-lower}
\end{equation}

\paragraph{Proof:}
If $Z^{(q)}\neq Z^{(k)}$, then there must exist some dimension $j$ such that $D^{(q)}_j\neq D^{(k)}_j$;
otherwise, by the packing definition~\eqref{eq:pack-L} we would have $Z^{(q)}=Z^{(k)}$, a contradiction.
Applying the union bound over $j$ and then Lemma~\ref{lem:one-digit-bound} yields the result.

\subsubsection{Lemma~\label{lem:mono}: Monotonicity}
The function $(L-1)/\log L$ is strictly increasing over $L\ge2$.

\paragraph{Proof:}
Let $g(L)=(L-1)/\log L$. Differentiating yields
$g'(L)=(\log L-1+1/L)/(\log L)^2$,
which is positive for $L>1$. 

Since $K=L^M$, we have $M=\log K/\log L$. Substituting into~\eqref{eq:stab-lower} gives
\begin{equation}
\mathrm{Stab}(L,M)\ge
1-2\delta f_{\max}\log K\cdot \frac{L-1}{\log L}.
\label{eq:stab-budget}
\end{equation}
Because $\log K$ is fixed, maximizing the lower bound on stability is equivalent to minimizing $(L-1)/\log L$. By Lemma~\ref{lem:mono}, this quantity is strictly increasing for $L\ge2$, and thus among all discretization schemes satisfying $K=L^M$, choosing $L=2$ yields the largest stability lower bound.

\subsection{Spurious Collision Rate Analysis}

We next analyze the lower bound of the spurious collision rate under a fixed vocabulary size $K$.

\subsubsection{Lemma~\label{lem:coll-lb}: A lower bound on collision probability}
Let a discrete symbol $Z$ have support size at most $K$ with marginal distribution $p(z)$. Then
\begin{equation}
\mathrm{Coll}(p)\ge \frac{1}{K},
\label{eq:coll-lb}
\end{equation}
with equality if and only if $p(z)=1/K$.

\paragraph{Proof:}
By the Cauchy--Schwarz inequality,
$\sum_z p(z)^2\ge (\sum_z p(z))^2/K=1/K$. 

\subsubsection{Lemma~\label{lem:binary-achieve}: Balanced binary discretization nearly achieves the bound}
If in binary discretization each digit satisfies
$\Pr[b=1]=\Pr[b=0]=1/2$,
and the digits are approximately independent in the marginal sense, then the packed symbol
$Z\in\{0,\dots,2^M-1\}$ is approximately uniformly distributed, and hence
$\mathrm{Coll}(p)\approx 1/K$.

\paragraph{Proof:}
Under the above conditions, each bit-string occurs with probability $(1/2)^M=1/K$, so the symbol distribution is approximately uniform. 

\subsection{Proof of Theorem~1}

\paragraph{Proof:}
Under a fixed vocabulary budget $K$, take any $L\ge2$ and set $K=L^M$.

By~\eqref{eq:stab-budget} and Lemma~\ref{lem:mono}, the stability lower bound is maximized at $L=2$, so binary discretization provides the strongest worst-case guarantee on hit stability.

On the other hand, by Lemma~\ref{lem:coll-lb}, the spurious collision rate of any discretization scheme is lower bounded by $1/K$; by Lemma~\ref{lem:binary-achieve}, balanced binary discretization can attain this bound.

Therefore, under a given vocabulary budget $K$, binary discretization therefore offers the strongest worst-case stability guarantee while achieving the minimum collision lower bound under the stated conditions.

\section{Quantized Attention and ROSA}

This section provides the full derivation of Theorem~2. For clarity, we first consider the computation of causal self-attention at time step $t$ under a single-batch, single-route (i.e., single-head) setting; we then explain the correspondence between this form and the ROSA retrieval mechanism proposed in this paper (Equation~\eqref{eq:dest_time}).

\subsection{Single-step form of causal self-attention}

Under the causal masking constraint, the attention output at time step $t$ can be written as
\begin{align}
\mathbf{o}_t
&=\sum_{i=0}^{t-1}\alpha_{t,i}\,\mathbf{v}_i, \label{eq:attn_out}\\
\alpha_{t,i}
&=\frac{\exp(\beta\, s_{t,i})}{\sum_{j=0}^{t-1}\exp(\beta\, s_{t,j})}, \label{eq:attn_softmax}
\end{align}
where $\mathbf{v}_i$ is the value vector at position $i$, $s_{t,i}$ denotes the similarity score between the query and the key, and $\beta>0$ is a scaling factor (equivalently, the inverse of the softmax temperature: a lower temperature corresponds to a larger $\beta$). Due to causality, the summation ranges only over historical positions $i<t$.

\subsection{$0$--$1$ match similarity and the extreme-preference regime}

Theorem~2 considers an extreme degenerate setting in which the similarity function performs only a $0$--$1$ ``match/mismatch'' test. Specifically, let
\begin{equation}
s_{t,i}
=\mathbb{I}\!\left[\text{key}(i)\ \text{matches query}(t)\right]
\in\{0,1\},
\label{eq:01_score}
\end{equation}
where $\mathbb{I}[\cdot]$ is the indicator function. In discrete-symbol modeling, $\text{key}(i)$ and $\text{query}(t)$ can correspond to a single symbol, or to an encoding of a context substring. The ROSA mechanism in this paper leverages a suffix automaton to maintain the matching relation of whether some suffix of the current query appears in the historical key string (see \S3.3).

Let the set of matched positions be
\begin{equation}
\mathcal{M}_t=\{\, i\in\{0,\dots,t-1\}\mid s_{t,i}=1 \,\},
\qquad
m_t=|\mathcal{M}_t|.
\label{eq:match_set}
\end{equation}
Substituting Equation~\eqref{eq:01_score} into the softmax definition in Equation~\eqref{eq:attn_softmax} yields an explicit form of the attention weights:
\begin{equation}
\alpha_{t,i}
=
\begin{cases}
\dfrac{e^{\beta}}{m_t e^{\beta} + (t-m_t)}, & i\in\mathcal{M}_t,\\[0.8em]
\dfrac{1}{m_t e^{\beta} + (t-m_t)}, & i\notin\mathcal{M}_t.
\end{cases}
\label{eq:alpha_closed_form}
\end{equation}

As $\beta\to\infty$, softmax exhibits an extreme preference for matched items. As long as $m_t>0$, i.e., there exists at least one matched position, we have
\begin{equation}
\lim_{\beta\to\infty}\alpha_{t,i}
=
\begin{cases}
\dfrac{1}{m_t}, & i\in\mathcal{M}_t,\\[0.4em]
0, & i\notin\mathcal{M}_t.
\end{cases}
\label{eq:alpha_limit}
\end{equation}
Substituting Equation~\eqref{eq:alpha_limit} back into the attention output in Equation~\eqref{eq:attn_out}, we obtain
\begin{equation}
\lim_{\beta\to\infty}\mathbf{o}_t
=\frac{1}{m_t}\sum_{i\in\mathcal{M}_t}\mathbf{v}_i.
\label{eq:uniform_avg}
\end{equation}
This shows that when the similarity function degenerates to a $0$--$1$ match test and the normalization process has an extreme preference for matched items, attention no longer learns continuous weights, but instead computes an equally weighted average over the value vectors at all matched positions. The conclusion of Theorem~2 follows directly from Equation~\eqref{eq:uniform_avg}.

\subsection{Correspondence to ROSA: from matches to reading successor values}

The ROSA retrieval in \S3.3 does not return the matched positions themselves; rather, it returns the successor time step of the end position of the most recent occurrence of the matched substring (see Equation~\eqref{eq:dest_time}):
\[
\tau_t=
\begin{cases}
\mathrm{endpos}(s_t)+1, & \text{if this position exists and is}<t,\\
-1, & \text{otherwise.}
\end{cases}
\]
This operation is equivalent to reading the successor values associated with the ``matched end-position set,'' i.e., taking $\mathbf{v}_{i+1}$ from position $i+1$.

To align exactly with the form in Equation~\eqref{eq:uniform_avg}, it suffices to view the value sequence in attention as already shifted to successor positions, i.e., define $\tilde{\mathbf{v}}_i=\mathbf{v}_{i+1}$ (and ignore out-of-range terms). Then the limiting attention output can be written as
\begin{equation}
\lim_{\beta\to\infty}\tilde{\mathbf{o}}_t
=\frac{1}{m_t}\sum_{i\in\mathcal{M}_t}\mathbf{v}_{i+1},
\label{eq:uniform_avg_successor}
\end{equation}
which matches the retrieval semantics of ROSA as ``jumping from the current context to a relevant historical continuation'': the match relation determines the candidate set $\mathcal{M}_t$, while the output is read from the successor values of these matched positions.

In implementation, the model first obtains run-level $\mathrm{endpos}$ on the symbol sequence via the suffix automaton, then maps it back to the original time axis to obtain the successor time index $\tau_{b,r,t}$, and reads the corresponding route value $a^{(v)}$ from that time step. The read-out is then unpacked and injected into the continuous representation space (Equations~\eqref{eq:unpack}--\eqref{eq:inj}). Therefore, ROSA can be viewed as implementing, in the discrete symbol space, the match-driven global read-out described by Equation~\eqref{eq:uniform_avg_successor}, and further performing continuous fusion via trainable injection parameters together with local sliding-window attention.

\subsection{Effects of the multi-route structure and RLE}

The above derivation holds for the single-route case. For the multi-route structure, one only needs to define a match set $\mathcal{M}_{t,r}$ for each route and perform the same uniform aggregation or successor read-out, and then concatenate the per-route read-outs or linearly project them back to $\mathbb{R}^C$; this does not change the basic form of the derivation.

Moreover, the RLE mechanism folds consecutive identical symbols into runs and updates the matching state only at run boundaries; in essence, it compresses the indexing of the candidate set. Under the ``match/mismatch'' semantics, the successor-position set obtained by mapping run-level matches back to the time axis remains consistent with the match structure on the original sequence, and therefore does not affect the conclusions in Equations~\eqref{eq:alpha_limit}--\eqref{eq:uniform_avg_successor}.

\section{Backpropagation and Counterfactual Gradient Derivation}
\label{app:cfgrad}

This section provides the complete derivations of the gradient formulas used in Section~\ref{sec:cfgrad}. To simplify notation and the derivation, we omit the temperature scaling term throughout, and uniformly use
$\sigma(x)=\frac{1}{1+e^{-x}}$ and its derivative $\sigma'(x)=\sigma(x)(1-\sigma(x))$.

\subsection{Sources of Non-differentiability}
\label{app:cfgrad:non-diff}

Recall the forward computation: ROSA's injection operation is determined by the following discrete chain:
\[
\begin{aligned}
\mathbf{Q}^{\mathrm{vec}},\mathbf{K}^{\mathrm{vec}},\mathbf{V}^{\mathrm{vec}}
&\xrightarrow{\;\mathbb{I}[\cdot>0]\;}
\mathbf{Q}^{\mathrm{bit}},\mathbf{K}^{\mathrm{bit}},\mathbf{V}^{\mathrm{bit}} \\[4pt]
&\xrightarrow{\;\mathrm{pack}\;}
\mathbf{a}^{(q)},\mathbf{a}^{(k)},\mathbf{a}^{(v)} \\[4pt]
&\xrightarrow{\;\mathrm{dest\_time}\;}
\tau \\[4pt]
&\xrightarrow{\;\mathrm{read\&unpack}\;}
\hat{\mathbf{b}} \\[4pt]
&\xrightarrow{\;e_0,e_1,\mathbf{W}_{\mathrm{out}}\;}
\mathrm{inj}.
\end{aligned}
\]

Here, $\mathrm{dest\_time}$ is produced deterministically by the SAM over the symbol sequence, and the threshold function $\mathbb{I}[\cdot>0]$ is a prototypical non-differentiable operator.
Therefore, $\mathrm{inj}$ is a piecewise-constant function of $(\mathbf{Q}^{\mathrm{vec}},\mathbf{K}^{\mathrm{vec}},\mathbf{V}^{\mathrm{vec}})$, which makes direct backpropagation numerically unstable and can even fail entirely.

To obtain stable and usable gradients, ROSA-Tuning adopts a counterfactual differentiation strategy:
for each query/key bit, we precompute the counterfactual retrieval indices when that bit is forcibly set to $0$ or $1$, thereby expressing the influence of a single bit on the loss as the difference between the read-outs of two counterfactual branches.

To simplify the subsequent notation, let $\tau_{b,r,t}$ denote the retrieval destination in the true forward pass (see Equation~\eqref{eq:dest_time}),
and define the validity mask
$m_{b,r,t}=\mathbb{I}[\tau_{b,r,t}\ge 0]$ (see Equation~\eqref{eq:mask}).
We also adopt the indexing convention $c\leftrightarrow(r,j)$, where $j\in\{0,\dots,M-1\}$.

\subsection{Intermediate Quantities}
\label{app:cfgrad:theta}

From Equation~\eqref{eq:inj}, we define the gradient of the injection vector as
\[
\mathbf{G}^{\mathrm{inj}}_{b,t,:} \triangleq \frac{\partial \mathcal{L}}{\partial \mathrm{inj}_{b,t,:}},
\qquad
\mathbf{G}^{y}_{b,t,:}
=
\frac{\partial \mathcal{L}}{\partial y_{b,t,:}}
=
\mathbf{W}_{\mathrm{out}}^{\top}\mathbf{G}^{\mathrm{inj}}_{b,t,:}.
\]

According to Equation~\eqref{eq:y}, the effective residual of $y$ at each dimension $c$ naturally contains
$\Delta_c=e_{1,c}-e_{0,c}$.
We therefore introduce the following contraction coefficient:
\begin{equation}
\theta_{b,t,c} \triangleq G^{y}_{b,t,c}\,\Delta_c.
\tag{\ref{eq:theta_def}}
\end{equation}

All subsequent derivations of the gradients with respect to $(\mathbf{q},\mathbf{k},\mathbf{v})$ can be uniformly expressed as inner products or aggregation forms between $\theta$ and the corresponding counterfactual read-out differences.

\subsection{Gradients w.r.t.\ $(e_0,e_1)$ and $\mathbf{W}_{\mathrm{out}}$}
\label{app:cfgrad:e01-wout}

To avoid confusion with the batch index $b$, this subsection uses $\hat{b}_{b,t,c}$ to denote the retrieved bit (corresponding to $b_{b,t,c}$ in Equation~\eqref{eq:unpack} in the main text).

From Equation~\eqref{eq:y}, $y$ can be rewritten as
\[
y_{b,t,c}
=
m_{b,r(c),t}
\Big((1-\hat{b}_{b,t,c})e_{0,c}+\hat{b}_{b,t,c}e_{1,c}\Big).
\]
Thus,
\[
\frac{\partial y_{b,t,c}}{\partial e_{0,c}} = m_{b,r(c),t}(1-\hat{b}_{b,t,c}),
\qquad
\frac{\partial y_{b,t,c}}{\partial e_{1,c}} = m_{b,r(c),t}\hat{b}_{b,t,c}.
\]

Multiplying both sides by $G^y_{b,t,c}=\partial\mathcal{L}/\partial y_{b,t,c}$,
and summing over $(b,t)$ yields
\begin{equation}
\begin{aligned}
\frac{\partial \mathcal{L}}{\partial e_{0,c}}
&=
\sum_{b,t}
m_{b,r(c),t}\bigl(1-\hat{b}_{b,t,c}\bigr)\,G^y_{b,t,c}, \\[4pt]
\frac{\partial \mathcal{L}}{\partial e_{1,c}}
&=
\sum_{b,t}
m_{b,r(c),t}\hat{b}_{b,t,c}\,G^y_{b,t,c}.
\end{aligned}
\label{eq:grad_e01}
\end{equation}

On the other hand, from $\mathrm{inj}_{b,t,:}=\mathbf{W}_{\mathrm{out}}y_{b,t,:}$ we obtain
\begin{equation}
\frac{\partial \mathcal{L}}{\partial \mathbf{W}_{\mathrm{out}}}
=
\sum_{b,t}
\mathbf{G}^{\mathrm{inj}}_{b,t,:}\,y_{b,t,:}^{\top}.
\label{eq:grad_wout}
\end{equation}

\subsection{Gradients w.r.t.\ $\mathbf{V}^{\mathrm{vec}}$: Destination-Scatter Aggregation}
\label{app:cfgrad:v}

In backpropagation we use the continuous surrogate
$P^{(v)}_{b,t,c}=\sigma(v^{\mathrm{vec}}_{b,t,c})$.
When $\tau_{b,r,t}\ge 0$, the read-out for the $m$-th bit dimension of route $r$ (with $c=(r,m)$) is
\[
\hat{b}_{b,t,(r,m)} = P^{(v)}_{b,\tau_{b,r,t},(r,m)}.
\]
When $\tau_{b,r,t}=-1$, we have $m_{b,r,t}=0$ and the injection for this route is identically zero,
so we can write uniformly
\[
\hat{b}_{b,t,(r,m)} = m_{b,r,t}\,P^{(v)}_{b,\tau_{b,r,t},(r,m)}.
\]

From Equation~\eqref{eq:y}, we have
\[
\frac{\partial \mathcal{L}}{\partial \hat{b}_{b,t,c}} = \theta_{b,t,c}.
\]
Together with
\[
\frac{\partial \hat{b}_{b,t,c}}{\partial P^{(v)}_{b,\tau,c}}
=
m_{b,r,t}\,\mathbb{I}[\tau_{b,r,t}=\tau],
\qquad
\frac{\partial P^{(v)}_{b,\tau,c}}{\partial v^{\mathrm{vec}}_{b,\tau,c}}
=
\sigma'(v^{\mathrm{vec}}_{b,\tau,c}),
\]
the chain rule gives
\[
\frac{\partial \mathcal{L}}{\partial v^{\mathrm{vec}}_{b,\tau,c}}
=
\sigma'(v^{\mathrm{vec}}_{b,\tau,c})
\sum_{t=0}^{T-1}
\theta_{b,t,c}\,
m_{b,r(c),t}\,
\mathbb{I}[\tau_{b,r(c),t}=\tau].
\]
Note that when $\tau_{b,r,t}=-1$, we have $m_{b,r,t}=0$, and the corresponding term vanishes automatically.
Removing the redundant mask yields Equation~\eqref{eq:grad_v_main} in the main text.

\subsection{Gradients w.r.t.\ $\mathbf{Q}^{\mathrm{vec}}$: Bitwise Counterfactual Differencing}
\label{app:cfgrad:q}

Fix batch $b$, time $t$, route $r$, and bit $j$. Let
$a^{(q)}_{b,t,r}$ denote the packed query symbol in the true forward pass (Equation~\eqref{eq:pack_q}).
Define the counterfactual symbol $a^{(q,u)}_{b,t,r}$ by forcing the $j$-th bit to $u\in\{0,1\}$, and perform one matching update on the same SAM state (determined by the history and the current prefix) to obtain the counterfactual destination
$\tau^{(u)}_{b,t,r,j}\in\{-1,0,\dots,t-1\}$.
In implementation, we precompute $\tau^{(0)},\tau^{(1)}$ on the CPU per query run and then map them back to per-time-step indices.

For any bit dimension $m$ within the same route (with $c=(r,m)$), we define the counterfactual read-out as
\[
\hat{b}^{(u)}_{b,t,(r,m)}
= m^{(u)}_{b,t,r,j}\,
  P^{(v)}_{b,\tau^{(u)}_{b,t,r,j},(r,m)},
\qquad u\in\{0,1\},
\]
where $m^{(u)}_{b,t,r,j}=\mathbb{I}[\tau^{(u)}_{b,t,r,j}\ge 0]$,
and define the difference
\[
\delta P^{(v)}_{b,t,r,m}(j)
\triangleq
\hat{b}^{(1)}_{b,t,(r,m)}-\hat{b}^{(0)}_{b,t,(r,m)}.
\]

Let
$s^{(q)}_{b,t,(r,j)}\triangleq \sigma(q^{\mathrm{vec}}_{b,t,(r,j)})$.
By expressing ``the effect of the $j$-th query bit on the read-out'' as a linear interpolation between the two counterfactual branches, its derivative with respect to $s^{(q)}$ is
\[
\frac{\partial \hat{b}_{b,t,(r,m)}}
     {\partial s^{(q)}_{b,t,(r,j)}}
= \delta P^{(v)}_{b,t,r,m}(j).
\]

Therefore,
\[
\begin{aligned}
\frac{\partial \mathcal{L}}
     {\partial s^{(q)}_{b,t,(r,j)}}
&= \sum_{m=0}^{M-1}
   \frac{\partial \mathcal{L}}
        {\partial \hat{b}_{b,t,(r,m)}}
   \frac{\partial \hat{b}_{b,t,(r,m)}}
        {\partial s^{(q)}_{b,t,(r,j)}} \\[6pt]
&= \sum_{m=0}^{M-1}
   \theta_{b,t,(r,m)}\,
   \delta P^{(v)}_{b,t,r,m}(j).
\end{aligned}
\]

Using
$\frac{\partial s^{(q)}}{\partial q^{\mathrm{vec}}}
 = \sigma'(q^{\mathrm{vec}})$,
we recover Equation~\eqref{eq:grad_q_main} in the main text:
\[
\frac{\partial \mathcal{L}}
     {\partial q^{\mathrm{vec}}_{b,t,(r,j)}}
=
\sigma'\!\big(q^{\mathrm{vec}}_{b,t,(r,j)}\big)\,
\sum_{m=0}^{M-1}
\theta_{b,t,(r,m)}\,
\delta P^{(v)}_{b,t,r,m}(j).
\]

\subsection{Gradients w.r.t.\ $\mathbf{K}^{\mathrm{vec}}$: Run-level Surrogate and Aggregation}
\label{app:cfgrad:k}

For each route, the key sequence is first folded by RLE, and the SAM then runs over the resulting run-level symbol sequence.
Let $\ell$ denote the run index, and let $\mathrm{start}(\ell)$ denote the start position of the $\ell$-th run on the original time axis.

In backpropagation, we allow only the continuous logits of keys at run starts to participate in gradient computation, and define
\[
u_{b,\ell,r,j}
\triangleq
\sigma\!\big(k^{\mathrm{vec}}_{b,\mathrm{start}(\ell),(r,j)}\big).
\]
Meanwhile, we define a continuous surrogate of values at the same run starts as
\[
P^{(v)}_{b,\ell,(r,m)}
\triangleq
\sigma\!\big(v^{\mathrm{vec}}_{b,\mathrm{start}(\ell),(r,m)}\big).
\]
For $k^{\mathrm{vec}}$ at non-run-start positions, we ignore its higher-order influence on the folding boundaries and the retrieval structure, and set its gradient to $0$.

For each time step $t$, route $r$, and query bit $j$ within this route,
as in Appendix~\S\ref{app:cfgrad:q}, we precompute the run-level destination indices of the two query-bit counterfactual branches,
$r\_idx^{(0)}_{b,t,r,j}$ and $r\_idx^{(1)}_{b,t,r,j}$;
if no valid hit exists, the corresponding index is set to $-1$.
When differentiating with respect to keys, we treat these two candidate indices as constants, i.e., we do not differentiate through their dependence on $k$,
thereby avoiding the substantial computational cost of explicitly modeling how flipping key bits changes the candidate set.

To make key learning differentiable, we define the following run-level surrogate.
Fix $(b,t,r,j)$ and any $m\in\{0,\dots,M-1\}$, and let
\[
\ell^{(0)} \triangleq r\_idx^{(0)}_{b,t,r,j},\qquad
\ell^{(1)} \triangleq r\_idx^{(1)}_{b,t,r,j}.
\]
By convention, when $\ell^{(u)}=-1$ the corresponding contribution is $0$ (equivalently, the mask is $0$).
The surrogate read-out for dimension $m$ in the route induced by query bit $j$ is defined as
\begin{equation}
\tilde{b}^{(k)}_{b,t,(r,m)}(j)
\triangleq
u_{b,\ell^{(1)},r,j}\,P^{(v)}_{b,\ell^{(1)},(r,m)}
-
u_{b,\ell^{(0)},r,j}\,P^{(v)}_{b,\ell^{(0)},(r,m)}.
\label{eq:k_surrogate_readout}
\end{equation}
The surrogate is intended to assign differentiable credit only between the two candidate runs from the query counterfactuals, rather than modeling how ``flipping key bits changes the candidate set.''

Since $\partial \mathcal{L}/\partial \hat{b}_{b,t,(r,m)}=\theta_{b,t,(r,m)}$ (see Equation~\eqref{eq:theta_def}),
we define the surrogate objective for the key branch as
\begin{equation}
\tilde{\mathcal{L}}_{k}
\triangleq
\sum_{b,t,r}\sum_{j=0}^{M-1}\sum_{m=0}^{M-1}
\theta_{b,t,(r,m)}\,
\tilde{b}^{(k)}_{b,t,(r,m)}(j).
\label{eq:k_surrogate_obj}
\end{equation}

For any $(b,\ell,r,j)$, differentiating Equation~\eqref{eq:k_surrogate_obj} yields
\[
\begin{aligned}
\frac{\partial \tilde{\mathcal{L}}_k}{\partial u_{b,\ell,r,j}}
&=
\sum_{t=0}^{T-1}\sum_{m=0}^{M-1}
\theta_{b,t,(r,m)}\,
P^{(v)}_{b,\ell,(r,m)} \\
&\quad\times
\Big(
\mathbb{I}\!\left[r^{(1)}_{\text{idx},\,b,t,r,j}=\ell\right]
-
\mathbb{I}\!\left[r^{(0)}_{\text{idx},\,b,t,r,j}=\ell\right]
\Big).
\end{aligned}
\]

Accordingly, we introduce the run-level accumulators
\begin{equation}
\begin{aligned}
U^{(1)}_{b,\ell,r,j}
&\triangleq
\sum_{t=0}^{T-1}\sum_{m=0}^{M-1}
\theta_{b,t,(r,m)}\,
P^{(v)}_{b,\ell,(r,m)}\,
\mathbb{I}[r\_idx^{(1)}_{b,t,r,j}=\ell],\\
U^{(0)}_{b,\ell,r,j}
&\triangleq
\sum_{t=0}^{T-1}\sum_{m=0}^{M-1}
\theta_{b,t,(r,m)}\,
P^{(v)}_{b,\ell,(r,m)}\,
\mathbb{I}[r\_idx^{(0)}_{b,t,r,j}=\ell],
\end{aligned}
\label{eq:U01_def}
\end{equation}
so that
\[
\frac{\partial \tilde{\mathcal{L}}_k}{\partial u_{b,\ell,r,j}}
=
U^{(1)}_{b,\ell,r,j}-U^{(0)}_{b,\ell,r,j}.
\]
Combining $u=\sigma(k^{\mathrm{vec}})$, we finally obtain
\begin{equation}
\frac{\partial \tilde{\mathcal{L}}_k}{\partial k^{\mathrm{vec}}_{b,\mathrm{start}(\ell),(r,j)}}
=
\sigma'\!\big(k^{\mathrm{vec}}_{b,\mathrm{start}(\ell),(r,j)}\big)\,
\Big(U^{(1)}_{b,\ell,r,j}-U^{(0)}_{b,\ell,r,j}\Big),
\label{eq:grad_k_main}
\end{equation}
which is exactly the same as Equation~\eqref{eq:grad_k_main} in the main text.
In implementation, this gradient is scattered back to the original time positions via the run-start index array.

\subsection{Gradients w.r.t.\ Projection Matrices and Gating: Standard Backpropagation}
\label{app:cfgrad:proj-gate}

From Equations~(9)--(11) (i.e., the definitions of $\mathbf{Q}^{\mathrm{vec}},\mathbf{K}^{\mathrm{vec}},\mathbf{V}^{\mathrm{vec}}$), we have
\[
\begin{aligned}
\mathbf{Q}^{\mathrm{vec}} &= \mathbf{U}\mathbf{W}_q, \quad
\mathbf{K}^{\mathrm{vec}} = \mathbf{U}\mathbf{W}_k, \\
\mathbf{V}^{\mathrm{vec}} &= \mathbf{U}\mathbf{W}_v, \quad
\mathbf{U} = \mathrm{LN}(\mathbf{H}).
\end{aligned}
\]

Therefore, after obtaining
$\partial\mathcal{L}/\partial\mathbf{Q}^{\mathrm{vec}}$,
$\partial\mathcal{L}/\partial\mathbf{K}^{\mathrm{vec}}$,
and
$\partial\mathcal{L}/\partial\mathbf{V}^{\mathrm{vec}}$,
the gradients of the three projection matrices can be computed directly using the standard backpropagation formulas for linear layers.
Similarly, the pre-attention mixing
$\mathbf{M}=(1-\boldsymbol{\alpha})\mathbf{H}+\boldsymbol{\alpha}\mathrm{inj}$
and the post-attention additive fusion operation are both differentiable operators, and their gradient computation requires no special handling.

\section{Additional Experiments}
\label{sec:appendix_exp}

This section provides two sets of experimental results that are directly related to the main conclusions.
The first set is on the MQAR task, which validates the direct gains of ROSA-Tuning in long-sequence retrieval and alignment;
the second set is an ablation study on the discrete symbol width $M$, which motivates the choice of our default hyperparameter setting.

\subsection{MQAR Experiments}
\label{sec:appd:mqar}

MQAR~\citep{arora2023zoologymeasuringimprovingrecall} is commonly used to evaluate a model's ability to recall information that appeared earlier in the given context. Prior work has shown that performance on MQAR reflects a model's in-context learning and information retrieval capability; as a result, it has become an important benchmark for evaluating language model architecture designs.

In our experiments, we set the sequence length to 512 and the window size to $W=32$, so that Window-Attn can hardly perform cross-segment retrieval using only local attention. Under the same training setup, we compare the validation accuracy of Global-Attn, Window-Attn, and ROSA + Window-Attn with model dimension $=128$. As shown in Table~\ref{tab:mqar_acc}, ROSA + Window-Attn reaches close to or equal to 100\% validation accuracy as early as epochs 4--5; both its convergence speed and final performance are substantially better than models using only Global-Attn or Window-Attn. In particular, Window-Attn is almost unable to learn this task, while Global-Attn gradually improves accuracy but converges noticeably more slowly overall. These results indicate that ROSA significantly enhances the model's ability for multi-item retrieval and match-based alignment under long-sequence settings.

\begin{table}[t]
\centering
\small
\caption{MQAR}
\begin{tabular}{cccc}
\toprule
\textbf{Epoch} & \textbf{Global-Attn} & \textbf{Window-Attn ($W{=}32$)} & \textbf{ROSA + Window-Attn} \\
\midrule
4 & 1.8  & 2.2 & 99.6 \\
5 & 22.4 & 2.6 & \textbf{100.0} \\
6 & 44.6 & 2.0 & \textbf{100.0} \\
7 & 61.2 & 3.0 & \textbf{100.0} \\
\bottomrule
\end{tabular}
\label{tab:mqar_acc}
\end{table}

\subsection{Ablation on ROSA Symbol Width}
\label{sec:appd:ablation_M}

ROSA's discrete symbols are formed by combining $M$ binary bits within each route, resulting in an alphabet size of $K = 2^M$. Increasing $M$ improves the expressivity of ROSA, but also increases the number of SAM transition branches and the computational cost of updating the matching states. This subsection analyzes the effect of different alphabet sizes on model performance, and motivates a reasonable default choice.

We perform ROSA-Tuning on Qwen3-0.6B, using PG19-train for training and PG19-test~\citep{raecompressive2019} for evaluation. We freeze all backbone model parameters and train only the newly introduced ROSA-Tuning parameters. As shown in Table~\ref{tab:M}, the test perplexity (PPL) exhibits a slight increasing trend as $M$ grows. Considering performance, computational efficiency, and generalization, we use $M=4$ as the default in all other experiments.

\begin{table}[t]
\centering
\small
\begin{minipage}{0.48\linewidth}
  \centering
  \caption{Ablation results for the discrete symbol width $M$}
  \begin{tabular}{cc}
    \toprule
    $M$ & Test PPL \\
    \midrule
    2 & 19.62 \\
    4 & 19.63 \\
    6 & 19.72 \\
    8 & 19.78 \\
    \bottomrule
  \end{tabular}
  \label{tab:M}
\end{minipage}
\hfill
\begin{minipage}{0.48\linewidth}
  \centering
  \caption{post-attn vs.\ pre-attn}
  \begin{tabular}{cc}
    \toprule
    Method & Test PPL \\
    \midrule
    pre-attn & 19.60 \\
    post-attn & 19.63 \\
    \bottomrule
  \end{tabular}
  \label{tab:pre}
\end{minipage}
\end{table}

\subsection{post-attn vs.\ pre-attn}

Under the same experimental setup as in Section~\ref{sec:appd:ablation_M}, we compare two ROSA fusion schemes. As shown in Table~\ref{tab:pre}, pre-attn achieves slightly lower test perplexity than post-attn, suggesting that fusing the injection term earlier typically yields a modest performance gain.

From an engineering perspective, post-attn can overlap with attention computation via a CPU--GPU pipeline, whereas pre-attn requires $\mathrm{inj}$ to be available before attention can run. Therefore, we adopt post-attn by default in the main experiments to balance overall efficiency; if an application prioritizes peak performance, pre-attn may be preferred.

\end{document}